\documentclass[conference]{IEEEtran}
\usepackage{cite}
\usepackage{amsmath,amssymb,amsfonts}
\usepackage{algorithmic}
\usepackage{graphicx}
\usepackage{textcomp}
\usepackage{xcolor}
\def\BibTeX{{\rm B\kern-.05em{\sc i\kern-.025em b}\kern-.08em
    T\kern-.1667em\lower.7ex\hbox{E}\kern-.125emX}}

\usepackage{subcaption}

\ifCLASSINFOpdf
\else
\fi
\begin{document}

\title{VP-SLAM: A Monocular Real-time Visual SLAM with Points, Lines and Vanishing Points\\
}

\author{\IEEEauthorblockN{Andreas Georgis}
\IEEEauthorblockA{\textit{School of ECE, NTUA} \\
Athens, Greece \\
georgisandreas@gmail.com}
\and
\IEEEauthorblockN{Panagiotis Mermigkas}
\IEEEauthorblockA{\textit{School of ECE, NTUA} \\
Athens, Greece \\
p.mermigkas@gmail.com
}
\and
\IEEEauthorblockN{Prof. Petros Maragos}
\IEEEauthorblockA{\textit{School of ECE, NTUA} \\
Athens, Greece \\
maragos@cs.ntua.gr
}
}


%


\maketitle

\begin{abstract}
Traditional monocular Visual Simultaneous Localization and Mapping (vSLAM) systems can be divided into three categories: those that use features, those that rely on the image itself, and hybrid models. In the case of feature-based methods, new research has evolved to incorporate more information from their environment using geometric primitives beyond points, such as lines and planes. This is because in many environments, which are man-made environments, characterized as "Manhattan world", geometric primitives such as lines and planes occupy most of the space in the environment. The exploitation of these schemes can lead to the introduction of algorithms capable of optimizing the trajectory of a Visual SLAM system and also helping to construct an exuberant map. Thus, we present a real-time monocular Visual SLAM system that incorporates real-time methods for line and VP extraction, as well as two strategies that exploit vanishing points to estimate the robot's translation and improve its rotation.Particularly, we build on ORB-SLAM2, which is considered the current state-of-the-art solution in terms of both accuracy and efficiency, and extend its formulation to handle lines and VPs to create two strategies the first optimize the rotation and the second refine the translation part from the known rotation. First, we extract VPs using a real-time method and use them for a global rotation optimization strategy. Second, we present a translation estimation method that takes advantage of last-stage rotation optimization to model a linear system. Finally, we evaluate our system on the TUM RGB-D benchmark and demonstrate that the proposed system achieves state-of-the-art results and runs in real time, and its performance remains close to the original ORB-SLAM2 system.
\end{abstract}



%
\IEEEpeerreviewmaketitle

\section{Introduction}
Visual SLAM (vSLAM) systems try to estimate a robot's location based on the multi-view geometry of the scene, combined with computer vision algorithms, while generating a 3D map of the environment. It is a critical tool for 3D reconstruction, image refinement, 3D holographic application, visual place recognition, AR/VR reality, and autonomous vehicles like micro air vehicles (MAVs). Various vSLAM approaches have been created based on various sensors such as single camera, stereo camera, RGB-D camera and event-camera. Moreover, feature-based approaches have traditionally attracted the most attention, since, they rely on well-suited computer vision algorithms to extract features and are more resistant to changes in light than direct methods. However, in low-textured or man-made environments where the extracted point features are not well-distributed or sufficient, incorporating other geometric primitives from multi-view geometry into the system, such as lines planes, or VPs (Vanishing Points) can boost the robustness of these systems.  \cite{b1}, \cite{b2}, \cite{b3}, \cite{b4}. 

On the other hand, most applications in practice work on certain scenarios, such as man-made environments.In these environments, to boost the performance of the vSLAM system, the hypothesis of the Manhattan World (\emph{MW}) is used \cite{b5}. The \emph{MW}, is a man-made environment with significant structural regularity, and with most areas of surrounding environment being described as a box world with three mutual orthogonal dominant directions. As a result, each \emph{MW} is associated with a frame, which is denoted as Manhattan Frame (\emph{MF}) and can be inferred from the VPs, which is the intersection of the image projections of the 3D parallel lines in \emph{MW}. As a result, employing the VPs can lead to a reduction in pose drift \cite{b6}, \cite{b7}.

Motivated by the insights made above, we present a vSLAM system that integrates simple computer vision algorithms for extracting lines, and VPs to reduce the drift in pose and optimize it.  More specifically, the main contributions of this paper are summarized below:: 
\begin{itemize}
\item Employ real-time computer vision algorithms for extract-
ing lines and VPs..
\item An optimization strategy of absolute rotation based on VPs.
\item A simple linear system for estimate translation.
\end{itemize}


\section{Related Work}
Next we briefly review the related works on vSLAM system with the focus on features and on leveraging the structural regularity to improve system performance. First, we have ORB-SLAM2 \cite{b8} which is a popular feature-based monocular vSLAM system, that extends the multi-threaded and keyframe-based architecture of PTAM \cite{b9}. It uses ORB features, builds a co-visibility graph and performs loop closing and localization tasks.To improve the robustness of point-based methods, the authors in \cite{b9} extracted lines from the environment and propose an algorithm to integrate them into a monocular Extended Kalman Filter SLAM system (EKF-SLAM). Finally, in PL-SLAM \cite{b1} points and lines extracted concurrently into a point-based system.

On the other hand, we have vSLAM systems based on the \emph{MW} assumption, like \cite{b6} which proposes a 2 stage \emph{MF} tracking to estimate the rotation of the pose based on the VPs and parallel lines and a refinement strategy to pose optimization. \cite{b10} proposes a mean-shift algorithm to track the rotation of \emph{MF} across scenes, while using 1-D density alignments for translation estimation. OPVO \cite{b11} improves the translation estimation by using the KLT tracker. Both methods require two planes to be visible in the frame at all times. LPVO \cite{b12} eases this requirement by incorporating lines into the system. Structural lines are aligned with the axes of the dominant \emph{MF} and can be integrated into the mean shift algorithm, improving robustness. Hence, for LPVO, only a single plane is required to be visible in the scene, given the presence of lines. Drift can still occur in translation estimation, as it relies on frame-to-frame tracking. 


\section{System Overview}
The proposed vSLAM system, addresses three major challenges: real-time complexity, optimize global rotation utilizing orthogonality and parallelism constraints, and a simple translation refinement using the information of global rotation. So, in the following paragraphs, each module of the proposed VP-SLAM system is presented briefly.

\begin{figure}[h]
\centerline{\includegraphics[scale=0.2]{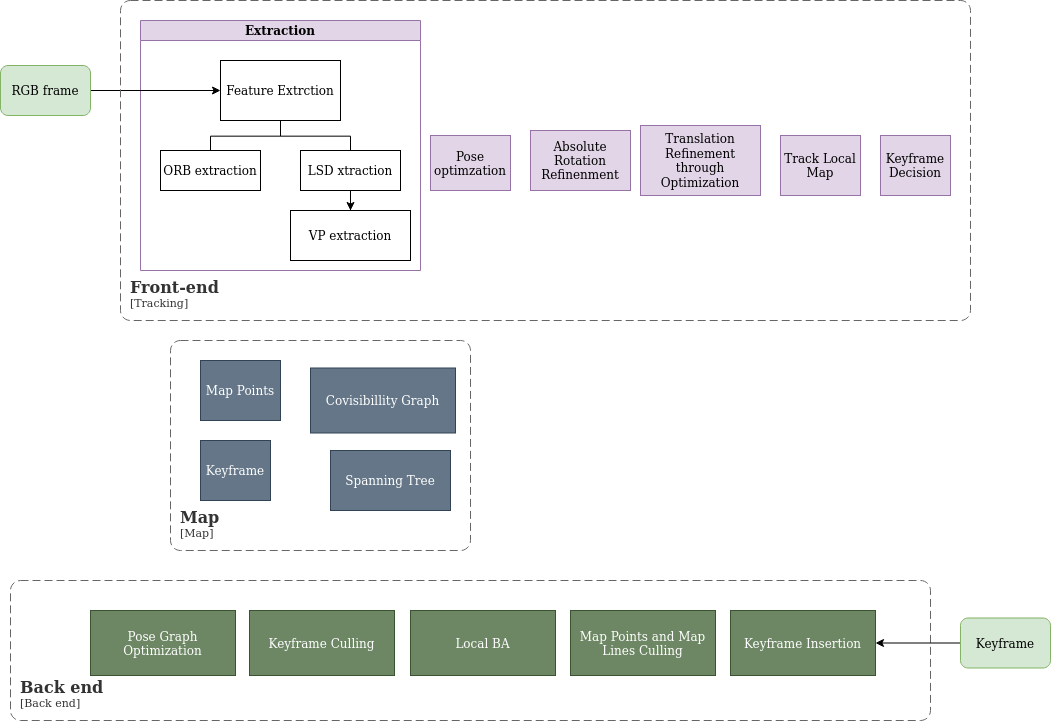}}
\caption{VP-SLAM pipeline.}
\label{fig1}
\end{figure}

The proposed VP-SLAM system is illustrated in ``Fig.~\ref{fig1}'', which is based on the ORB-SLAM2 \cite{b8} architecture. The system is composed of two modules: the front-end and the back-end. The front-end is responsible for real-time VO, providing ego-motion estimation, with local optimization for the current frame and keyframe decision, while the back-end is responsible for the map representation, local map optimization, and the strategy for inserting and culling keyframes.

Similar to \cite{b1}, in the front-end section, points and lines are output in parallel in each RGB frame. Then, similar to ORB-SLAM2 \cite{b8}, we use the constant velocity motion model to obtain an initial pose estimation, and then use the points and lines to optimize it. Then, to further optimize the rotation, we present an optimization approach that incorporates the extracted VP and the information about the parallelism of the lines. Then, knowing the optimized rotation, we can refine the translation by solving a linear system, as shown in \cite{b6}. Finally, note that throughout the text, tables and vectors are in bold capital letters, while scalar gradients are represented as simple letters. Also, to represent the camera pose of the $i$-frame relative to the global coordinate system, we use the notation ( $\mathbf{R_{iw}}, \mathbf{t_{iw}}$ ) where $\mathbf{R_{iw}} \in \emph{SO(3)}$ is the rotation matrix and $\mathbf{t_{iw}} \in \mathbb{R}^{3xn}$ the translation vector.

We also use $\mathbf{\delta_k^i}$ to represent the set of directions of VPs in the $C_i$ frame, and with $\mathbf{d_k}$ as the set of three orthogonal dominant directions in \emph{MW}. Additionally, we define ``$[\cdot]_x$`` as a 3 x 3 skew symmetric matrix, and we use ''$\sim$'' to represent an equality up to a scale.

\subsection{Real time Complexity}\label{AA}
The computational complexity is one of the most important concerns to address in vSLAM algorithms. To  preserve the real-time properties of ORB-SLAM2 \cite{b8}, we carefully selected, and  utilized, fast methods for operating line extraction and VPs estimation.


\subsubsection{Lines}
To detect line segments in the image, we use the robust LSD (Line Segment Detector) detector \cite{b13} which is an \emph{O(n)} line segment detector where $n$ is the number of pixels in the image. To describe, the line segments we use the LBD (Line Band Descriptor) descriptor \cite{b14} which relies on local appearance of lines and geometric constraints while preserving the real time complexity and robustness against image artifacts.


\subsubsection{VPs Estimation}
To extract the VPs we adopt the \cite{b15} method, which is based on 2 lines and guarantees that the execution time is \emph{O(n)} and the extracted vanishing points are optimal and orthogonal VPs in \emph{MW} . The main idea of the method is to exploit the Gaussian sphere as the parameter space of the rotation, with the principal point $(x_0,y_0)$\textsuperscript{T} being the center of the sphere. Thus, two parallel lines in 3D are projected onto the Gaussian sphere as two great circles that intersect at a point. The direction of this point from the origin of the sphere is considered as a candidate vanishing point direction ( $v_{1}$ ). To achieve real-time performance, a polar grid is created that intersects the image plane and spans the latitude and longitude of half of the Gaussian sphere with a size of 90x360 and a precision of 1\textsuperscript{o}. Thus, a pair of lines intersecting at a point in the image plane contributes to the corresponding cell of the polar grid by weight:
\begin{equation*}
\emph{score}=\emph{score} + ||l_0||\times ||l_1|| \times \sin(2\theta)  \label{1} 
\end{equation*}
where $||l_0||$ and $||l_1||$ represent the length of two line segments in pixel and $\theta$ is the angle between them and \emph{score} is the accumulated score on each polar grid cell.
After creating the first candidates VPs ($v_{1}$) directions and considering the orthogonal constraint, the second VP ($v_{2}$) must lie on the great circle that is perpendicular to the great circle of ($v_{1}$) generating 360 evenly distributed candidates VPs $({v_i\textsuperscript{2}}_{i=1}^{360})$ directions. Given the first ($v_{1}$) and the second ($v_{2}$) VPs directions, the third candidate VPs directions can be obtained by the cross product of each pair of $v_{1}$ and $({v_i\textsuperscript{2}}_{i=1}^{360})$.Thus, given all candidate directions of VPs, the best estimate of VPs are calculated from the set (VPs) with the highest score, where the score of a (VPs) hypothesis set is the sum of the scores of three polar grid cells that belong to the three associated VP directions.


\subsection{Absolute Rotation Optimization}
Inspired by \cite{b6}, \cite{b7} we propose a strategy that further optimizes the camera rotation $C_{i}$ by exploiting the extracted VPs and the set of three mutually orthogonal dominant directions $\mathbf{d_k }$ = [$\mathbf{d_i}^{1}$ $|$ $\mathbf{d_i}^{2}$ $|$ $\mathbf{d_{i}}^{3}$] of the Manhattan world (\ emph{MW}) where $\mathbf{d_i}^{j}$ is a column vector. Specifically, given an image with a cluster of 3D parallel lines in the scene, these lines must be aligned with a dominant direction $\mathbf{d_k^i}$ in \emph{MW}. Thus, given at least two clusters of lines in the image, the normal vector of the great circle on the Gaussian sphere of the corresponding line in the associated cluster $\mathbf{s_i}$ must be perpendicular to the dominant direction of the cluster. Therefore, for the $i$-th camera frame $C_{i}$ , we can formulate a least-square problem to calculate the set of dominant directions $\mathbf{d_k^i}$ in the current frame as below:
\begin{equation}
    \mathbf{S}^T \mathbf{d_k^i}=0 \label{2}
\end{equation}
where $\mathbb{S} \in \mathbb{R}^{3xn}$ and n is the number of lines in the cluster. The normals  $\mathbf{s_i}$ represent the columns of $\mathbf{S}$ can be obtained from $\mathbf{s_i}$ = $\mathbf{K^T}$ $\mathbf{l_i}$ where $\mathbf{K}$ is the camera matrix and $\mathbf{l_i}$ = $(\mathbf{sp_i} \times \mathbf{ep_i} / ||\mathbf{sp_i} \times \mathbf{ep_i}||)$ where $\mathbf{sp_i},\mathbf{ep_i} \in \mathbb{R}^{2}$ are the endpoints of the line on the image plane. Finally, solving this problem with SVD we obtain the set of mutual orthogonal dominant directions $\mathbf{d_k^i}$ in the current $C_{i}$ frame. Also, since, the direction of the VPs reflect the orientation of the current $C_{i}$ frame w.r.t \emph{MW}, and the initial set $\mathbf{d_k}$ computed from the Eq.~\eqref{2} (instead of computed from VPs on the inital frame as in \cite{b7}) present the orientation of the initial frame $C_0$ w.r.t \emph{MW} we can construct the following equation that connects the estimated VP with the set of initial mutual orthogonal directions $\mathbf{d_k}$ as follow: 
\begin{equation*}
\mathbf{vp_k^i} \sim \mathbf{K} \mathbf{R_{iw}} \mathbf{d_k} \label{3}
\end{equation*}
Thus, as a result, we conclude that:
\begin{equation}
    \mathbf{\delta_k^i} \sim \mathbf{d_k^i} \sim \mathbf{R_{iw}} \mathbf{d_k} \label{4}
\end{equation}
where, $\mathbf{\delta_k^i}$ is the Vanishing Direction (VD) of the VP $\mathbf{vp_k^i}$.
\begin{equation*}
    \mathbf{\delta_k^i} = \mathbf{K^{-1}} \mathbf{vp_k^i}
\end{equation*}

Therefore, to optimize further the absolute rotation $\mathbf{R_{iw}}$ of the current $C_{i}$ frame with respect to condition Eq.~\eqref{4} we define the following cost function to minimize:
\begin{equation}
    \mathbf{E(\omega_i)} =  \sum_{k=1}^{3} \mathbf{E_k(\omega_i)} = \sum_{k=1}^{3} \arccos{(\mathbf{\delta_k^i}\cdot\mathbf{R_{iw}} \mathbf{d_k})} \label{5}
\end{equation}
Note that if the initial frame $C_{0}$ has not at least two clusters with enough lines, we continue with the next frame until we find a frame $C_{i}$ that meets the condition. Additionally, the $\omega_i$ in Lie algebra is mapping from $\mathbf{R_{iw}}$ in Lie group. We employ the Levenberg–Marquardt (LM) optimization to minimize the cost function $\mathbf{E(\omega_i)}$. The Jacobians of the cost function $\mathbf{E(\omega_i)}$ are:

\begin{equation*}
    \begin{split}
    \mathbf{J_k} & = -\frac{1}{\sqrt{1-(\delta_k^i\cdot\mathbf{d_k})^2}}\cdot \delta_k^i 
    \frac{\partial\mathbf{d_k}}{\partial w_i} \\  
   & = \frac{1}{\sqrt{1-(\delta_k^i\cdot\mathbf{d_k})^2}}\cdot \delta_k^i \cdot([\mathbf{d_k}]_x),
    \label{6}
    \end{split}
\end{equation*}
The initial value of $\mathbf{R_{iw}}$ is obtained by optimizing both the reprojection error of lines and reprojection error of points. 


\subsection{Translation Refinement}
Having optimized the global rotation $\mathbf{R_{iw}}$ from the previous step we can use this information to construct a linear system to solve for refine the translation. Moreover, the availability of global rotation $\mathbf{R_{iw}}$ can lead to simplify the initial non-linear reprojection error to a linear one. Specifically, from the:
\begin{equation*}
    \mathbf{E_p} =  \sum_{i=1}^{N} \rho(||\mathbf{x_i} - \pi(\mathbf{X_i},\mathbf{R_{iw}},\mathbf{t_{iw}},\mathbf{K})||^2) \label{10}
\end{equation*}
with known the absolute rotation, we lead to:
\begin{equation}
    \mathbf{t_{iw}}=((\mathbf{R_{iw}}\cdot \mathbf{X_i})^{(3)} + t^{(3)})\cdot
    \begin{bmatrix}
    \frac{u_i - c_x}{f_x}\\
   \frac{v_i - c_y}{f_y}
    \end{bmatrix}- 
    \begin{bmatrix}
    \mathbf{(R_{iw}}\cdot \mathbf{X_i})^{(1)} + t^{(1)}\\
   \mathbf{(R_{iw}}\cdot \mathbf{X_i})^{(2)} + t^{(2)}
    \end{bmatrix}
,
\label{11}
\end{equation}
where $[]^{(j)}$ is $j$-th row of a vector. Re-arranging \eqref{11} we lead to minimizing the following system:
\begin{equation}
    \mathbf{t_{iw}}=\min(||\mathbf{A}\cdot \mathbf{t} - \mathbf{b}||^2)
\label{12}
\end{equation}
where: 
\begin{equation*}
\mathbf{A} =\begin{bmatrix}
                    -1 && 0 &&  \frac{u_i - c_x}{f_x}\\
                     0 && -1 && \frac{v_i - c_y}{f_y}
                \end{bmatrix}
                \label{13}
\end{equation*}
and  
\begin{equation*}
\mathbf{b} = \begin{bmatrix}
                    (\mathbf{R_{iw}}\cdot \mathbf{X_i})^{(1)} - (\mathbf{R_{iw}}\cdot \mathbf{X_i})^{(3)} \cdot \frac{u_i - c_x}{f_x}\\
                       (\mathbf{R_{iw}}\cdot \mathbf{X_i})^{(2)} - (\mathbf{R_{iw}}\cdot \mathbf{X_i})^{(3)} \cdot \frac{v_i - c_y}{f_y}
                \end{bmatrix}
\end{equation*}
The system in Eq.~\eqref{12} is a least square problem that can be solved under a RANSAC framework in order to find the maximum inlier set from all pairs of observations of the current ${C_i}$ frame to solve the following normal equation:
\begin{equation*}
    \mathbf{A^T}\cdot\mathbf{A}\cdot{t} =\mathbf{A} \cdot \mathbf{b}
\end{equation*}



\section{Experiment Evaluation}
To validate the performance of the proposed SLAM system in terms of computation time and accuracy, we compare our SLAM system with the state-of-the art ORB-SLAM2 \cite{b8} in real-world scenarios using the TUM RGB-Benchmark \cite{b10}. All experiments were carried out with an AMD Ryzen 5 2600 6-core CPU. In this paper, we aim at demonstrating the effectiveness of exploiting VPs and dominant directions in a \emph{MW}, for pose optimization. For, a fair comparison, we disable the loop closing function from both systems. This is due to the fact that when the loop closing module is enabled, the two systems will converge to the same trajectory and have the same absolute pose error, so we will not see the results of our method.


\subsection{Localization Accuracy in the TUM RGB-D Benchmark}\label{SCM}
We test our method on TUM-RGB-D dataset \cite{b16} which consists of several real-world camera sequences which contains a variety of scenes, like cluttered areas and scenes containing varying degrees of structure and texture, with ground truth trajectories and images with (640x480) resolution, recorded at full frame rate (30 Hz). Table.\ref{table:1} shows the absolute translation RMSE (Root Mean Square Error) of the compared VO methods and SLAM methods, and the estimated trajectories of different sequences in TUM-RGBD benchmark \cite{b16}.

\begin{table}[!ht]
\centering
\caption{TUM-RGB-D results RMSE ATE (m)}
\begin{tabular}{ || c c c c ||}

 \hline
 TUM-RGB-D Sequence & Ours & ORB-SLAM2  & PL-SLAM\\
 \hline
 fr1/desk    & 0.032  & $\mathbf{0.020}$ & 0.032\\
 fr1/plant   & $\mathbf{0.020}$      & $\mathbf{0.020}$ & 0.097\\
 fr2/desk    & 0.012  & $\mathbf{0.009}$ & 0.012\\
 fr3/large-cabinet & $\mathbf{0.100}$ & 1.420 & $\mathbf{0.100}$ \\
 fr3/long-office-house-validation & 0.019 & $\mathbf{0.010}$ & 0.019 \\
 fr3/structure-texture-far & 0.010 & 0.012 & $\mathbf{0.008}$\\
 fr3/structure-texture-far-near   & $\mathbf{0.014}$ & 0.015 & 0.014 \\
 fr3/nostr-text-near-withLoop & 0.177 & $\mathbf{0.015}$ & 0.177 \\
 fr3/nostr-text-far & $\mathbf{0.029}$ & 0.048 & 0.029 \\ 
 \hline
\end{tabular}
\label{table:1}
\end{table}

 

\begin{figure}[h]
	\centering
	\begin{subfigure}[t]{0.48\linewidth}
        \includegraphics[width=1.0\linewidth]{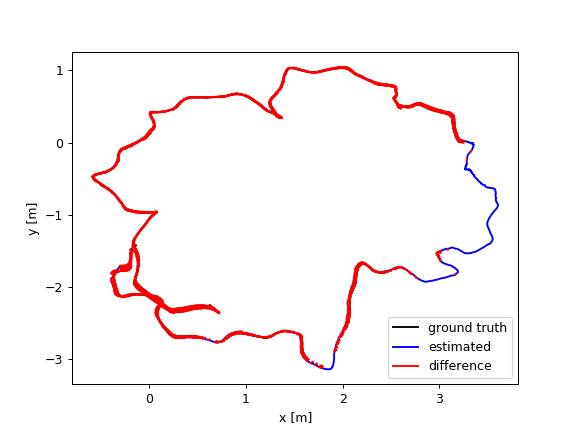}
	\end{subfigure}
	\begin{subfigure}[t]{0.48\linewidth}
		\includegraphics[width=1.0\linewidth]{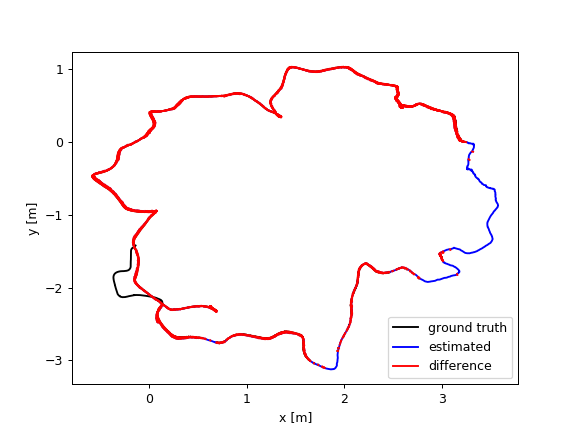}
	\end{subfigure}
	\caption{\small Estimated absolute trajectories and GT on TUM fr2/desk dataset for VP-SLAM (left) and ORB-SLAM2 (right) respectively.}
	\label{fig:traj}
	\vspace{-0.5cm}
\end{figure}

\begin{figure}[h]
\centerline{\includegraphics[scale=0.3]{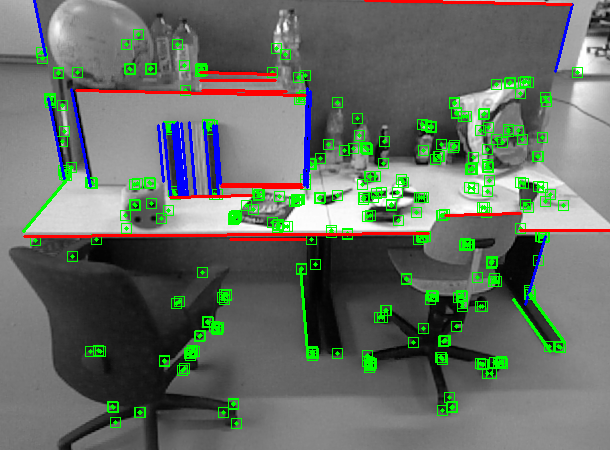}}
\caption{The three line clusters are shown in red, green and blue after VPs extraction with ORB-features in green rectangles.}
\label{fig4}
\end{figure}

All trajectories were aligned with 7DoF with the ground truth before computing the ATE (Absolute Translation Error) error with the script provided by the benchmark \cite{b16}. As shown in Table.\ref{table:1} even with disable the Loop Closure module, the accuracy of the proposed VP-SLAM, is near the state of the art. This is due to our concern of keeping the performance of VP-SLAM as real time as possible and concurrently utilizing additional information from the geometry of the environment. Moreover, we propose two simple programmable modules for optimization and refinement, compared to ORB-SLAM2 that uses non-linear pose graph optimizations at the back end. Additional, disabling the Loop Closure module the accumulating error of rotation can be alleviated to some extent, but as shown the trajectory optimized by our method is close to the ORB-SLAM2 \cite{b8}. Additionally, in Fig.~\ref{fig:traj}, (left) we see the trajectory that produces our VP-SLAM, in contrast with the ORB-SLAM2 Fig.~\ref{fig:traj} (right). Finally, in  Fig.~\ref{fig4} we present the line clusters that computed after extracting the VPs.

\begin{table}[htbp]
\caption{Time Execution of Tracking and Mapping}
\begin{tabular}{ |p{1.5cm}||p{1.5cm}|p{1.5cm}|p{1.5cm}|p{1.5cm}|  }
 \hline
 \multicolumn{5}{|c|}{Mean execution time in (ms)} \\
 \hline
 Thread & Operation & Ours & ORB-SLAM2 & PL-SLAM\\
 \hline
 \hline
 
            & Feature Extraction      & 96,94  &  15,79 & 96,64\\
            & VPs Extraction          & 46,56  &  ----  & ----\\
Tracking    & Initial Pose Estimation & 71,1   & 49,06  & 71,1\\
            & Track Local Map         & 29,93  & 26,61  & 29,93\\
 \hline
 \hline
&  Total    & 244,53 & \textbf{91,46} & 197,97 \\
\hline 
 \hline
 
                   & Keyframe Insertion  & 47,90   & 45,82   & 47,90 \\
                   & Map Feature Culling & 2,48    & 1,89    & 2,48\\
Local Mapping & Map feature Creation& 135,93  & 21,93   & 135,93\\
                   & Keyframe Culling    & 72,52   & 70,07   & 72,52\\
                   & Local BA            & 1499,37 & 1434,57 & 1499,37\\
         
\hline
\hline
&Total & 1758,2 & $\mathbf{1574,28}$ & 1711,64 \\
\hline 
\end{tabular}
\label{table:2}
\end{table}
 
 
         

\subsection{Time Complexity}
We also evaluate the runtime complexity of our VP-SLAM against the state of the art ORB-SLAM2 \cite{b8}. We summarize In Table.\ref{table:2} the time required for each subtask in ``Tracking`` and ``Local Mapping`` in our VP-SLAM and ORB-SLAM2 \cite{b8}. Note that the most time-consuming task is the local BA and the map feature creation in both systems. In conclusion, despite the line feature detection and VPs extraction combined with the optimization strategy increase the cost time, the whole system can still perform in real time.  


\section{Conclusion}
In this paper, we propose a real time monocular visual SLAM system that exploits the structure of man-made environments to further optimize the pose. More specifically, it is especially suitable for environments with more geometric structures, because it can detect VPs and lines from a single image. We propose two methods, one of which is leverage the information of VPs and the parallelism constraint of lines to construct an optimization strategy for rotation, while the second one uses the information from the former optimization to build a distinct refinement strategy for translation. Finally, the experiments on benchmark datasets with real world scenes show that the accuracy of the proposed system is close to the state of the art ORB-SLAM2 \cite{b8}. Additionally, the performance maintains real-time and indicate that drift could be further reduced.

In the future, we intend to strengthen our system with a stronger optimization technique and test it in more indoor environments.

\bibliographystyle{IEEEtran}

\begin{thebibliography}{99}
\bibitem{b1} A. Pumarola, A. Vakhitov, A. Agudo, A. Sanfeliu and F. Moreno-Noguer, "PL-SLAM: Real-time monocular visual SLAM with points and lines," 2017 IEEE International Conference on Robotics and Automation (ICRA), 2017, pp. 4503-4508, doi: 10.1109/ICRA.2017.7989522.
\bibitem{b2} Gomez-Ojeda, Ruben et al. “PL-SLAM: A Stereo SLAM System Through the Combination of Points and Line Segments.” IEEE Transactions on Robotics 35 (2019): 734-746.
\bibitem{b3} Hosseinzadeh, Mehdi et al. “Structure Aware SLAM Using Quadrics and Planes.” ACCV (2018).
\bibitem{b4} H. Li, J. Yao, J. -C. Bazin, X. Lu, Y. Xing and K. Liu, "A Monocular SLAM System Leveraging Structural Regularity in Manhattan World," 2018 IEEE International Conference on Robotics and Automation (ICRA), 2018, pp. 2518-2525, doi: 10.1109/ICRA.2018.8463165.
\bibitem{b5} J. Straub, N. Bhandari, J. J. Leonard and J. W. Fisher, "Real-time manhattan world rotation estimation in 3D," 2015 IEEE/RSJ International Conference on Intelligent Robots and Systems (IROS), 2015, pp. 1913-1920, doi: 10.1109/IROS.2015.7353628.
\bibitem{b6} J. Liu and Z. Meng, "Visual SLAM With Drift-Free Rotation Estimation in Manhattan World," in IEEE Robotics and Automation Letters, vol. 5, no. 4, pp. 6512-6519, Oct. 2020, doi: 10.1109/LRA.2020.3014648.
\bibitem{b7} H. Li, J. Yao, J. -C. Bazin, X. Lu, Y. Xing and K. Liu, "A Monocular SLAM System Leveraging Structural Regularity in Manhattan World," 2018 IEEE International Conference on Robotics and Automation (ICRA), 2018, pp. 2518-2525, doi: 10.1109/ICRA.2018.8463165.
\bibitem{b8} R. Mur-Artal, J. M. M. Montiel and J. D. Tardós, "ORB-SLAM: A Versatile and Accurate Monocular SLAM System," in IEEE Transactions on Robotics, vol. 31, no. 5, pp. 1147-1163, Oct. 2015, doi: 10.1109/TRO.2015.2463671.  
\bibitem{b9} G. Klein and D. Murray, “Parallel tracking and mapping for small AR workspaces,” in 2007 6th IEEE and ACM international symposium on mixed and augmented reality. IEEE, 2007, pp. 225–234.
\bibitem{b10} Zhou, Yi \& Kneip, Laurent \& Rodríguez, Cristian \& li, Hongdong. (2017). Divide and Conquer: Efficient Density-Based Tracking of 3D Sensors in Manhattan Worlds. 3-19. 10.1007/978-3-319-54193-81. 
\bibitem{b11} P. Kim, B. Coltin, and H. J. Kim, “Visual Odometry with Drift-free Rotation Estimation Using Indoor Scene Regularities,” in British Machine Vision Conference (BMVC), 2017
\bibitem{b12} P. Kim, B. Coltin, and H. J. Kim, “Low-drift visual odometry in structured environments by decoupling rotational and translational motion,” in 2018 IEEE international conference on Robotics and automation (ICRA). IEEE, 2018, pp. 7247–7253.
\bibitem{b13} R. Grompone von Gioi, J. Jakubowicz, J. -M. Morel and G. Randall, "LSD: A Fast Line Segment Detector with a False Detection Control," in IEEE Transactions on Pattern Analysis and Machine Intelligence, vol. 32, no. 4, pp. 722-732, April 2010, doi: 10.1109/TPAMI.2008.300.
\bibitem{b14} L. Zhang and R. Koch. An efficient and robust line segment matching approach based on LBD descriptor and pairwise geometric consistency. JVCIR, 24(7):794–805, 2013.
\bibitem{b15} X. Lu, J. Yaoy, H. Li, Y. Liu and X. Zhang, "2-Line Exhaustive Searching for Real-Time Vanishing Point Estimation in Manhattan World," 2017 IEEE Winter Conference on Applications of Computer Vision (WACV), 2017, pp. 345-353, doi: 10.1109/WACV.2017.45.
\bibitem{b16} J. Sturm, N. Engelhard, F. Endres, W. Burgard and D. Cremers, "A benchmark for the evaluation of RGB-D SLAM systems," 2012 IEEE/RSJ International Conference on Intelligent Robots and Systems, 2012, pp. 573-580, doi: 10.1109/IROS.2012.6385773.


\end{thebibliography}

\vspace{12pt}

\end{document}